\documentclass{article}
\usepackage{ijcai26}

\usepackage{times}
\usepackage{soul}
\usepackage{url}
\usepackage[hidelinks]{hyperref}
\usepackage[utf8]{inputenc}
\usepackage[small]{caption}
\usepackage{graphicx}
\usepackage{amsmath}
\usepackage{amsthm}
\usepackage{booktabs}
\usepackage{algorithm}
\usepackage{algorithmic}
\usepackage[switch]{lineno}

\usepackage{amssymb}
\usepackage{multirow}
\usepackage{bm}
\usepackage{makecell}
\usepackage{color}
\usepackage{bbding}
\usepackage{xcolor}
\usepackage{colortbl}
\usepackage{subcaption}
\usepackage{adjustbox}

\usepackage{hyperref}

\title{IPSM-Bench: A New Intermediate Phase Segmentation Benchmark in Microstructure Images of Zinc-Based Absorbable Biomaterials}

\author{
Jinglin Xu$^1$
\and
Shangyan Zhao$^4$\and
Jiabo Wang$^1$\and
Xinghong Mu$^1$\and
Yulong Lei$^1$\and
Jiacheng Zhang$^6$\and
Hongbo Sun$^3$\thanks{Corresponding Authors. Accepted by IJCAI 2026.}\And
Yageng Li$^{2,5*}$\\
\affiliations
$^1$School of Artificial Intelligence, University of Science and Technology Beijing, China\\
$^2$School of Advanced Materials Innovation, University of Science and Technology Beijing, China\\
$^3$China Telecom Artificial Intelligence Technology (Beijing) Co., Ltd\\
$^4$School of Materials Science and Engineering, University of Science and Technology Beijing, China\\
$^5$Institute of Materials Intelligent Technology, Liaoning Academy of Materials, China\\
$^6$School of Big Data and Software Engineering, Chongqing University, China
}

\begin{document}

\maketitle

\begin{abstract}
Zinc-based alloys are indispensable emerging absorbable metallic biomaterials, and their macroscopic performance is governed by microstructural characteristics. Intermediate phases—key microstructural constituents—are pivotal in regulating mechanical and functional properties. However, intermediate phase segmentation in zinc alloy microstructures faces formidable challenges: scarce annotated datasets, low contrast, difficulty detecting small targets, and heterogeneous morphologies. To this end, we construct IPSM-Bench, the largest high-quality dataset for zinc-alloy intermediate phase segmentation. Furthermore, we propose SCoP-SAM, a new Spatial Context Prior-guided SAM method that leverages the gradient structure and grayscale properties of intermediate phases to capture spatial context priors and incorporates them into the entire SAM encoding-decoding process, improving segmentation performance. Based on the proposed IPSM-Bench, we establish a new benchmark for intermediate phase segmentation to systematically evaluate state-of-the-art (SOTA) methods and advance research on zinc alloy microstructure analysis. Extensive experiments on IPSM-Bench and additional public alloy benchmarks demonstrate that our SCoP-SAM not only achieves SOTA performance for zinc-alloy intermediate phase segmentation but also generalizes remarkably well to other alloy scenarios.
\end{abstract}

\section{Introduction}

Zinc-based alloys (i.e., zinc alloys), an indispensable class of emerging and promising absorbable metallic biomaterials, have substantial application potential in orthopedics~\cite{huang2025additively}, cardiovascular medicine~\cite{yang2023lithium}, and dentistry~\cite{tong2023znp}. The microstructural characteristics of zinc alloys directly dictate their macroscopic performance, and intermediate phases—key microstructural components—play a pivotal role in regulating mechanical and functional properties. In real-world scenarios, the microstructure of zinc alloys exhibits complex multiphase characteristics; for example, Zn-Mg alloys comprise both lamellar eutectic morphologies and spherical precipitates. These intermediate phases span a broad size spectrum, exhibit uneven spatial dispersion, and display heterogeneous morphologies (e.g., lamellar, spherical, dendritic).

Traditional methods for representing intermediate phases primarily rely on techniques such as scanning electron microscopy (SEM) and optical microscopy (OM) to acquire microstructural images, followed by segmentation and statistical analysis utilizing manual annotations by materials professionals and simple image analysis software.
For one thing, manual annotation is inefficient and fails to meet the stringent requirements for accuracy and reproducibility. For another, existing image segmentation techniques face many challenges in zinc alloy microstructural analysis:
1) \textbf{\textit{Scarcity of Annotated Datasets}}: High-quality microstructural images of zinc alloys incur substantial acquisition costs and require precise annotation by materials science experts, resulting in a severe paucity of available training data.
2) \textbf{\textit{Low Contrast}}: The grayscale contrast between the matrix and intermediate phases in zinc alloys is often marginal, especially in SEM backscattered electron images, where the contrast between phases can differ by only a few grayscale levels.
3) \textbf{\textit{Difficulty in Small Target Detection}}: Zinc alloys contain a large number of submicron-scale intermediate phases, which occupy an extremely small pixel fraction in images, carry negligible semantic and textural information, and are highly susceptible to confusion with background textures.
4) \textbf{\textit{Challenges in Complex Morphologies}}: Intermediate phases in zinc alloys exhibit complex, heterogeneous morphologies, ranging from irregular forms (e.g., lamellar, spherical, dendritic, needle-like), often accompanied by overlap and agglomeration.
In summary, mitigating the scarcity of annotated datasets and accurately segmenting intermediate phases are critical prerequisites for addressing these challenges.

\begin{figure*}[t]
    \centering
\includegraphics[width=0.87\textwidth]{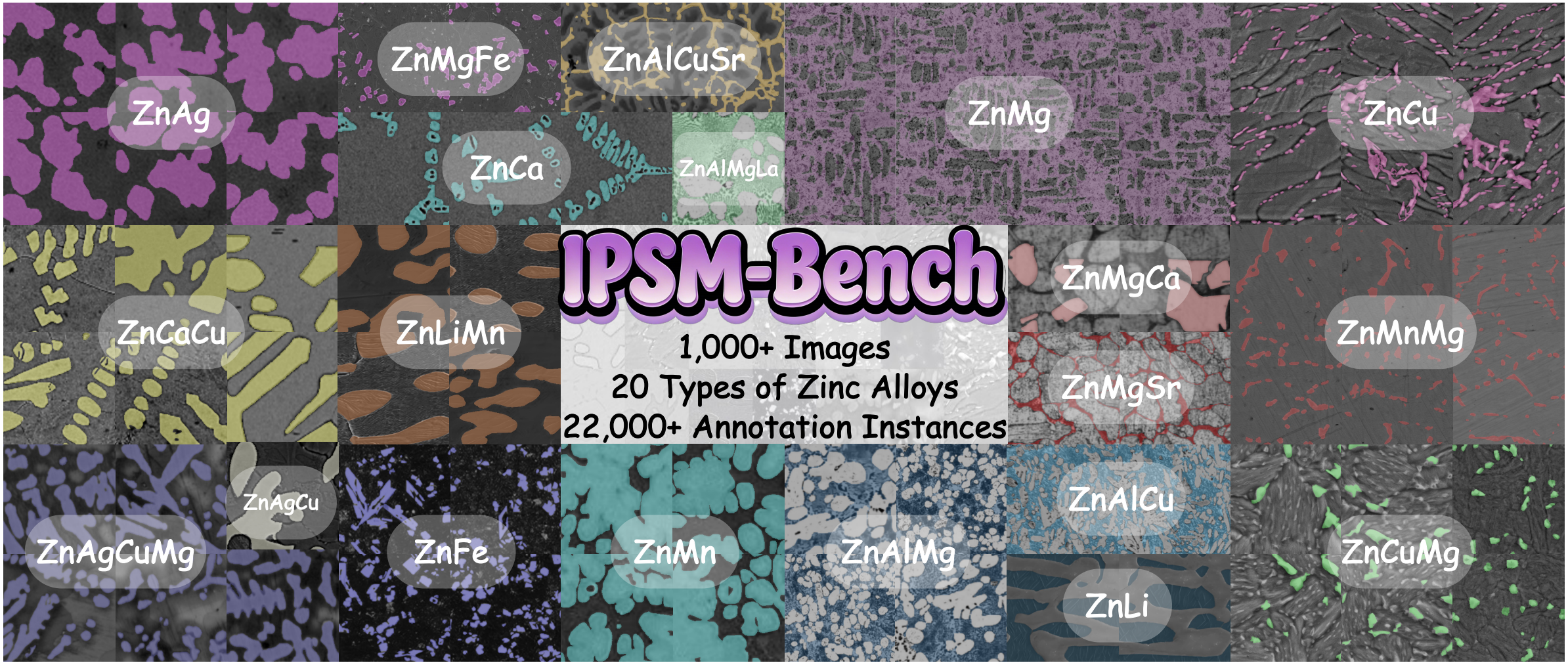}
    \caption{Overview of the proposed IPSM-Bench.}
    \label{top-benchmark}
\end{figure*}

\begin{table*}[t]
\centering
\adjustbox{width=\linewidth}{
      \setlength{\tabcolsep}{4pt}
\begin{tabular}{lllllcccc}
\toprule
\multirow{2}{*}{Dataset} & \multirow{2}{*}{Material} & \multirow{2}{*}{Available} & \multirow{2}{*}{Anno.Type} & \multirow{2}{*}{Resolution} & \multirow{2}{*}{Alloy Types} & Number of  & \multicolumn{2}{c}{Data Modality}             \\
\cmidrule(lr){8-9}
 & &  &  & &  & Images & SEM & OM \\ 
\midrule
NBS~\cite{stuckner2022microstructure}  & Ni-based Superalloy & Public   & Mask & 512$\times$512 & Unknown & 23 & \CheckmarkBold & \textcolor{lightgray}{\XSolidBrush} \\
UHCS-Seg~\cite{decost2019high}  & Ultrahigh Carbon Steel & Public   & Mask  & 645$\times$475 & 1 & 24 & \CheckmarkBold & \textcolor{lightgray}{\XSolidBrush} \\
DP590–1~\cite{li2025novel}  & Dual-phase Steel  & Private  & Mask  & 512$\times$512 & 1  & 23 & \CheckmarkBold & \textcolor{lightgray}{\XSolidBrush}  \\
DP590–2~\cite{li2025novel}  & Dual-phase Steel  & Private  & Mask & 1024$\times$768 & 1 & 21 & \CheckmarkBold & \textcolor{lightgray}{\XSolidBrush}  \\
Carbide~\cite{ma2023training} & Fe-0.2C-1.35Mn-2.5Cr-1.5Si Alloy  & Private  & Mask  & 2048$\times$1536 & 1  & 64 & \CheckmarkBold & \textcolor{lightgray}{\XSolidBrush}  \\
\textbf{IPSM-Bench}~(Ours) & \textbf{Zinc-based Alloys} & This Paper & \textbf{Mask} & \textbf{512$\times$512} & \textbf{20} & \textbf{1,054}  & \CheckmarkBold & \CheckmarkBold\\ 
\bottomrule
\end{tabular}
}
\caption{
Comparison of IPSM-Bench with existing intermediate phase segmentation datasets.
}
\label{tab:dataset_comparison}
\end{table*}

Therefore, we first construct a large-scale, high-quality annotated dataset of microstructural images of zinc alloys, named \textbf{\textit{IPSM-Bench}}\footnote{https://github.com/AgileMotionTeam/IPSM-Bench}, as shown in Figure~\ref{top-benchmark}. IPSM-Bench contains 1,054 microstructural images (512$\times$512), comprising SEM and OM images, covering 20 types of zinc-based alloys and 22,179 manually annotated intermediate phase instances. All annotations are performed by experts in metallic materials science under quality control, ensuring precise delineation of intermediate phase boundaries. This dataset addresses a critical gap in the field by providing scarce, high-quality annotated data that serves as a reliable foundation for training and evaluating intermediate phase segmentation models.
Furthermore, we propose a new \textbf{S}patial \textbf{Co}ntext \textbf{P}rior-guided SAM method, named \textbf{\textit{SCoP-SAM}}, which extracts intermediate phase gradient structure and grayscale property priors as spatial context and integrates them into both the encoder and decoder of the SAM (Segment Anything Model) to provide prior-aware prompts for intermediate phase regions. Our SCoP-SAM precisely anchors the target intermediate phase regions during segmentation and fully delineates their complex boundary contours, thereby significantly improving segmentation accuracy.

The main contributions are summarized as follows:
1) To the best of our knowledge, we construct the first large-scale, diverse, and extensively annotated high-quality benchmark \textbf{\textit{IPSM-Bench}} for intermediate phase segmentation in zinc alloy microstructures, providing a reliable foundation for training intermediate phase segmentation models and a fair comparison platform to drive methodological advances in analyzing zinc alloy microstructures.
2) We propose a new spatial context prior-guided SAM method (\textbf{\textit{SCoP-SAM}}) that incorporates gradient-structure and grayscale-property priors into SAM to address key challenges in intermediate phase segmentation—low contrast, difficulty detecting small targets, and complex morphologies.
3) Extensive experiments are conducted on the proposed IPSM-Bench benchmark dataset and two other representative public alloy datasets to demonstrate the effectiveness of our SCoP-SAM method for intermediate phase segmentation in microstructure images.

\begin{figure*}[t]
\centering
\includegraphics[width=0.87\textwidth]{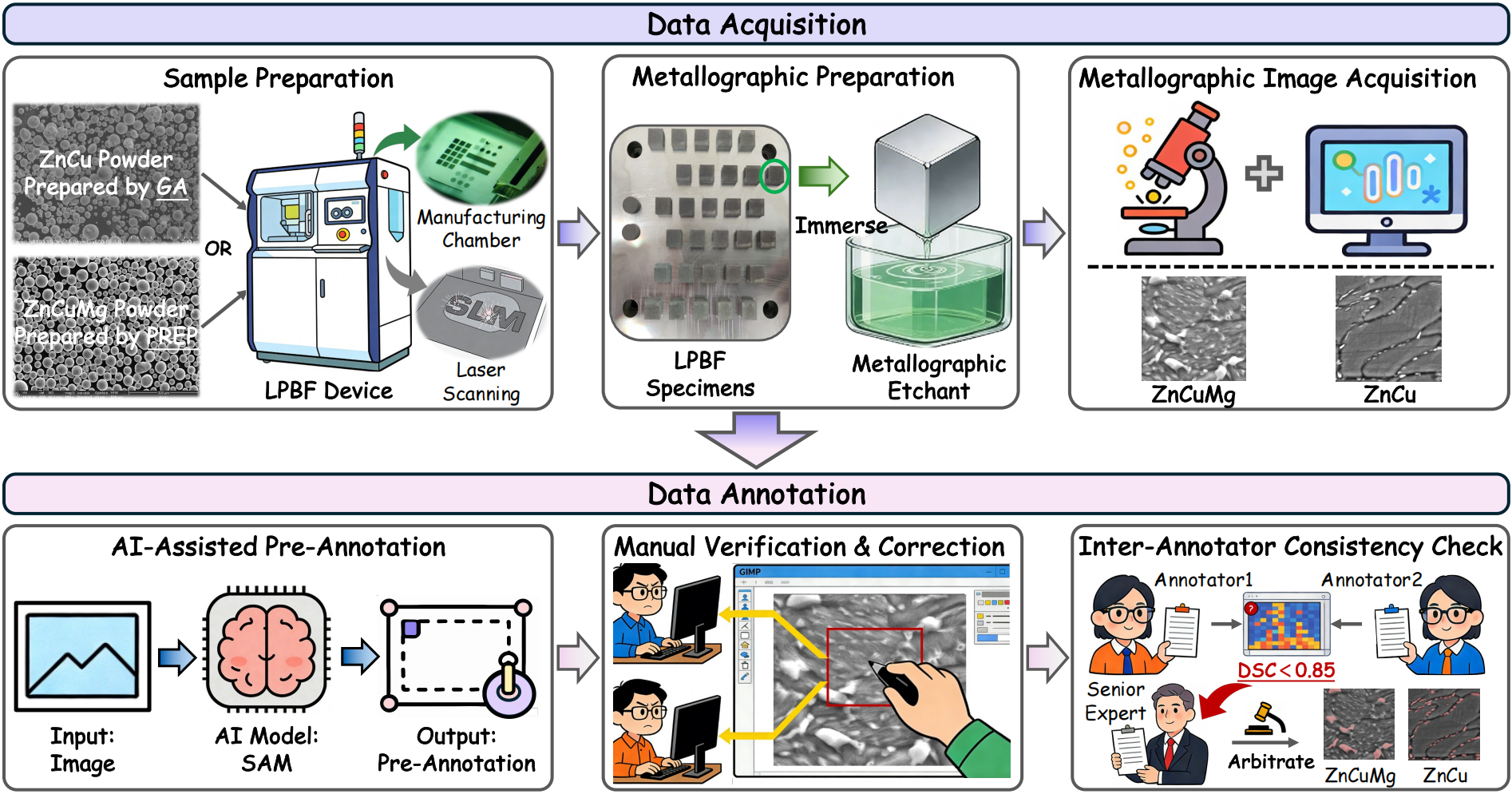}
\caption{Data acquisition and annotation pipeline for IPSM-Bench.}
\label{data-construct-pipeline}
\end{figure*}

\section{Related Work}

\subsection{Metallic Microstructure Image Datasets}

In the field of metallic microstructure image analysis, open-access, large, and finely annotated datasets for intermediate phases remain scarce.
As shown in Table~\ref{tab:dataset_comparison}, NBS~\cite{stuckner2022microstructure} comprised only 23 intermediate phase particle images of nickel-based superalloys, with three segmentation classes: matrix phase, secondary precipitates, and tertiary precipitates. UHCS~\cite{decost2017uhcsdb} focused on the pearlite-to-spheroidized cementite transformation in ultra-high carbon steel but lacked pixel-level annotations; its segmented variant, UHCS-Seg~\cite{decost2019high}, provided 24 images of intermediate phase particles. MetalDAM~\cite{luengo2022tutorial} contained 42 high-resolution SEM images of additively manufactured steel, with pixel-level annotations for five microstructural components; however, intermediate phases accounted for an extremely low proportion (about 0.24\%), leading to incomplete annotations that hindered the training of intermediate phase segmentation models.
Notably, recent work~\cite{li2025novel} has leveraged SAM for training-free microstructure analysis, but no standardized intermediate phase dataset has been established to date.
Thus, there is an urgent need to develop a larger, more diverse, and more thoroughly annotated high-quality dataset for training intermediate phase segmentation models and as a fair comparison platform to advance research in this field.

\subsection{SAM-based Image Segmentation Models in Multidisciplinary Applications}

SAM~\cite{kirillov2023segment}, a foundational segmentation model based on the prompt-learning paradigm, enables zero-shot segmentation and supports multiple prompt types, improving the accuracy of complex-structure analysis~\cite{ma2025alloy,abebe2025sam,guru2025machine,ma2024segment,marks2025cellsam,chen2024robustsam,wugausam,ji2024segment,wu2026hierarchical}.
Recently, SAM-Adapter~\cite{chen2023sam} developed a lightweight adaptation framework that converts task-specific knowledge into visual prompts to address poor generalization in specialized segmentation tasks.
CAT-SAM~\cite{xiao2024cat} proposed a few-shot conditional tuning network via a prompt-bridging structure that enables decoder-guided collaborative tuning of the encoder to address SAM's adaptation challenges.
SAMCT~\cite{lin2024samct} was equipped with a U-shaped CNN image encoder based on SAM, which provides supplementary local features for segmentation.
DenseSAM~\cite{zhou2025densesam} replaced SAM’s reliance on positional prompts in dense scenarios with semantic guidance, introducing an efficient semantic injection module and a dual-head decoding structure to tackle the challenge of dense objects in pathology and remote sensing images.
MatSAM~\cite{li2025novel} developed a SAM-based automatic analysis model for material microstructures by integrating a prompt-generation strategy and multi-scale feature fusion, enabling efficient segmentation of structures such as grains and second-phase particles.
$\mu$SAM~\cite{archit2025segment} provided a microscopy image segmentation and tracking tool based on SAM, unifying interactive and automatic segmentation across 2D/3D/temporal sequence data.
Despite SAM's efforts in microstructure analysis, existing methods still suffer from recognition accuracy for fine structures (e.g., grains, precipitates, microcracks) and a high susceptibility to misjudgment in complex phase-distribution scenarios.

\begin{figure*}[t]
\centering
\includegraphics[width=0.92\textwidth]{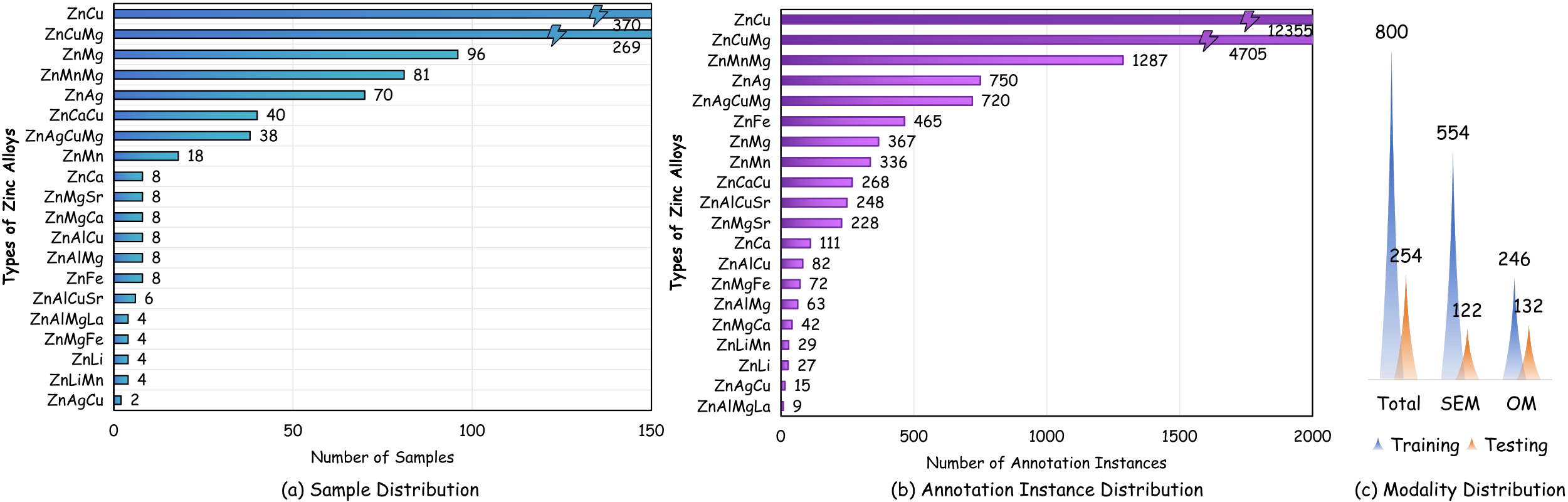}
\caption{Statistical information of IPSM-Bench.}
\label{statistics}
\end{figure*}

\section{Dataset: IPSM-Bench}

\subsection{Construction Necessity}
Zinc-based absorbable biomaterials exhibit unique advantages and broad application prospects in medical fields, owing to a moderate degradation rate matching the repair cycle of human bone tissue, excellent biocompatibility, and favorable mechanical properties~\cite{li2020additively}. Therefore, we construct a new microstructure image dataset, named \textbf{\textit{IPSM-Bench}}, spanning diverse zinc-based alloys to precisely analyze intermediate phases, which are critical for optimizing zinc alloy performance.

\subsection{Data Acquisition}
Our IPSM-Bench comprises 1,054 images, including 800 self-constructed and 254 sourced from the academic literature, covering two image types acquired via Optical Microscopy (OM) and Scanning Electron Microscopy (SEM). The data acquisition process is illustrated in Figure~\ref{data-construct-pipeline}, which includes sample preparation, metallographic preparation, and metallographic image acquisition.

\noindent\textbf{Sample Preparation.}
All samples were fabricated using the Laser Powder Bed Fusion (LPBF) technique. The raw materials were spherical zinc alloy powders produced by Gas Atomization (GA) and the Plasma Rotating Electrode Process (PREP). Fabrication was performed on a commercial LPBF system (SLM 125HL, SLM Solutions, Germany). Bulk solid samples were prepared by precisely adjusting laser processing parameters, with the laser energy density strictly maintained at 30-120 J/mm$^3$. Ultimately, bulk samples measuring 10$\times$10$\times$10mm were obtained. After removing the surface oxide layer, the samples' densities were measured using the Archimedes method. Only samples with densities above 99\% were selected for subsequent metallographic preparation.

\noindent\textbf{Metallographic Preparation.}
The surface parallel to the substrate was chosen for metallographic observation. Samples were subjected to a graded mechanical grinding process using 400\#, 800\#, 1500\#, 3000\#, and 5000\# abrasive papers in sequence (\#: grit of the sandpaper). After grinding, the surface was polished sequentially with W2.5$\mu$m, W1.5$\mu$m, and W0.5$\mu$m diamond metallographic polishing pastes until a bright, scratch-free surface suitable for microscopic observation was obtained. The polished observation surface was immersed in a dedicated enhanced metallographic etchant for 1-2 seconds, immediately removed, thoroughly rinsed with anhydrous ethanol, and dried prior to subsequent observation.

\noindent\textbf{Metallographic Image Acquisition.}
OM images of the microstructures are acquired using an optical microscope (BX53m, Olympus, Japan), while SEM images are obtained with a cold-field emission scanning electron microscope (SU8100, Hitachi, Japan).
In addition to the experimentally acquired images, literature-derived data were collected by downloading microstructure images reported in relevant academic literature. These images were further processed by removing interfering regions and retaining valid areas to ensure consistency and usability within the dataset.

\begin{figure*}[t]
\centering
\includegraphics[width=0.97\textwidth]{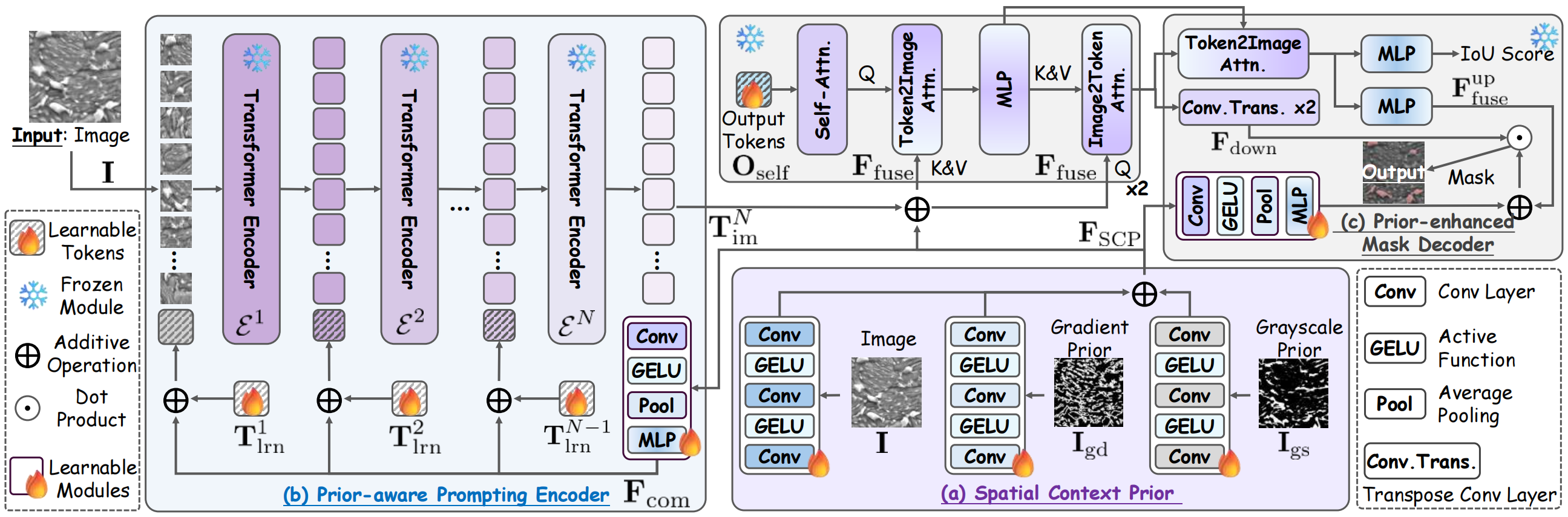}
\caption{Architecture of the proposed SCoP-SAM.
It comprises three core modules: (a) Spatial Context Prior, (b) Prior-aware Prompting Encoder, and (c) Prior-enhanced Mask Decoder.
}
\label{method-pipeline}
\end{figure*}

\subsection{Data Annotation}

To ensure annotation efficiency, we adopt an AI-assisted annotation strategy in this work. The entire annotation process comprises three key stages: AI model-based pre-annotation, manual verification and correction, and an inter-annotator consistency check.

\noindent\textbf{AI-Assisted Pre-Annotation.} A pre-trained segmentation model is used to generate initial segmentation masks for all images, thereby reducing the manual annotation workload. For zinc alloy intermediate phases with complex morphologies (e.g., lamellar, spherical, dendritic, and needle-like), SAM is selected as the AI auxiliary tool.

\noindent\textbf{Manual Verification and Correction.}
We invite two professional annotators with more than 5 years of experience analyzing the microstructures of metallic materials to manually verify and correct the AI-generated pre-annotations.
The correction criteria are based on the metallographic characteristics of zinc alloys: 1) \textit{Boundary Correction}: For intermediate phases with blurred or inaccurate segmentation boundaries, the annotators adjusted the mask edges using professional annotation software (GIMP) to ensure precise delineation of intermediate phase contours.
2) \textit{Missing and False Annotation Correction}: The annotators supplemented the missing small-scale intermediate phases that the AI model did not recognize and eliminated false annotations of matrix regions misidentified as intermediate phases.

\noindent\textbf{Inter-Annotator Consistency Check.}
To ensure the reliability of the annotated data, an inter-annotator consistency check was conducted after manual correction. The Dice Similarity Coefficient (DSC) is used as an evaluation metric to quantify the consistency between the two annotators' results. For annotations with DSC$<$0.85 (i.e., low consistency), a third senior materials science expert was invited to arbitrate and determine the final annotation. The average DSC of the final annotated dataset exceeds 0.90, indicating high annotation consistency and reliability.

\subsection{Statistics}\label{subsection:Statistics}
IPSM-Bench includes 20 Zinc-based alloy systems and comprises 1,054 images. Figure~\ref{statistics} (a) shows the distribution of Zinc alloy types, with ZnCu (370 samples, 35.10\% of the total) and ZnCuMg (269 samples, 25.52\% of the total) accounting for 60.63\% of the total. IPSM-Bench has 22179 annotation instances. Figure~\ref{statistics} (b) illustrates the distribution of annotations across Zinc alloy types, with ZnCu (12355 annotation instances, 55.71\% of the total) and ZnCuMg (4705 annotation instances, 21.21\% of the total) accounting for 76.92\% of the total.
Figure~\ref{statistics} (c) presents the distribution of data modalities across training and testing. The training set includes 554 SEM images and 246 OM images, while the test set contains 122 SEM images and 132 OM images.

\section{Methodology: SCoP-SAM}
In this work, we propose a new Spatial Context Prior-guided SAM method (\textbf{\textit{SCoP-SAM}}) for microstructure images.
As shown in Figure~\ref{method-pipeline}, our SCoP-SAM comprises three core modules: Spatial Context Prior, Prior-aware Prompting Encoder, and Prior-enhanced Mask Decoder.

\subsection{Spatial Context Prior (SCP)}

To construct a spatial context prior for each input image, the SCP module integrates three types of information—RGB image, gradient prior, and grayscale prior—via three weight-sharing-free learnable networks. Each network comprises three convolutional layers interleaved with two GELU nonlinearities.
Specifically, given an input image $\mathbf{I}$, we introduce three learnable networks $f_{\text{im}}$, $f_{\text{gd}}$, and $f_{\text{gs}}$ to compute the spatial context prior $\mathbf{F}_{\text{SCP}}$ as follows:
\begin{equation}
\mathbf{F}_{\text{SCP}}=f_{\text{im}}(\mathbf{I})+f_{\text{gd}}(\mathbf{I}_{\text{gd}})+f_{\text{gs}}(\mathbf{I}_{\text{gs}})
\end{equation}
where $\mathbf{F}_{\text{SCP}}$ has a dimensionality of $H_1\!\times\!W_1\!\times\!D_1$. Here, $\mathbf{I}_{\text{gd}}$ (gradient prior) and $\mathbf{I}_{\text{gs}}$ (grayscale prior) are derived via edge detection and grayscale thresholding algorithms, respectively.
For the gradient prior, we employ an adaptive edge-detection algorithm to generate a gradient structure map, integrating the Canny operator and the Otsu thresholding method to extract edge features. Then, by comparing contour counts, the edge map with a greater number of contours is selected as $\mathbf{I}_{\text{gd}}$, aiming to maximize the preservation of topological integrity and structural continuity of the intermediate phase edges.
For the grayscale prior, we model the grayscale distribution using dominant color analysis, yielding a grayscale map $\mathbf{I}_{\text{gs}}$. Specifically, we first analyze the distribution of black and white pixels in $\mathbf{I}_{\text{im}}$ and determine the target dominant color (black or white) based on the dominant pixel count. Subsequently, K-means clustering is applied to all image pixels, and the cluster center closest to the target dominant color is selected via Euclidean distance. Finally, a binary regional mask is generated around the cluster center, effectively capturing the aggregation properties of the intermediate phase region in the grayscale space.

\begin{table*}[t]
  \centering
  \adjustbox{width=\linewidth}{
    \setlength{\tabcolsep}{17pt}
    \begin{tabular}{llcccc}
      \toprule
      \multirow{2}{*}{Method} & \multirow{2}{*}{Venue} & \multirow{2}{*}{Backbone} &
      \multicolumn{3}{c}{IPSM-Bench (\%)}   \\
      \cmidrule(r){4-6}
      & & &
      mIoU  & mF1 & mPre   \\
      \midrule
      SAM~\cite{kirillov2023segment} & \textcolor{gray}{ICCV'23} & SAM ViT-H &29.05 &44.83 &60.98
      \\
      SAM-Adapter~\cite{chen2023sam} & \textcolor{gray}{ICCVW'23} & SAM ViT-H &55.12 &63.50 &70.49 \\
      SAMCT~\cite{lin2024samct}  & \textcolor{gray}{ICLR'24} & SAM ViT-H &
      \underline{62.01} & 69.14 &71.13  \\
      CAT-SAM~\cite{xiao2024cat} & \textcolor{gray}{ECCV'24} & SAM ViT-H &43.05 &55.60 &34.87  \\
      DenseSAM~\cite{zhou2025densesam} & \textcolor{gray}{IJCAI'25} & SAM ViT-H &59.82 &\underline{73.13} &\underline{84.79}  \\
      MatSAM~\cite{li2025novel} & \textcolor{gray}{Acta Materialia'25} & SAM ViT-H &29.01 &44.78 &61.02    \\
      $\mu$SAM ~\cite{archit2025segment}& \textcolor{gray}{Nature Methods'25} & SAM ViT-H &39.19 &44.08 &42.48 \\
      \rowcolor{blue!7}\textbf{SCoP-SAM} (Ours) & \textcolor{gray}{This Paper} & SAM ViT-H &
      \textbf{71.52} & \textbf{82.82} & \textbf{87.55}  \\
      \bottomrule
    \end{tabular}
  }
  \caption{Comparison results against state-of-the-art methods on IPSM-Bench. The best and second-best results are in \textbf{bold} and \underline{underline}.}
  \label{tab:compare-IPSM-Bench-multi-datasets} 
\end{table*}

\begin{table*}[t]
  \centering
  \adjustbox{width=\linewidth}{
    \setlength{\tabcolsep}{12pt}
    \begin{tabular}{llcccccc}
      \toprule
      \multirow{2}{*}{Method} & \multirow{2}{*}{Venue} &
      \multicolumn{3}{c}{NBS (\%)} & 
      \multicolumn{3}{c}{UHCS (\%)} \\
      \cmidrule(lr){3-5} \cmidrule(lr){6-8}
      & &
      mIoU & mF1 & mPre &
      mIoU & mF1 & mPre  \\
      \midrule
      SAM~\cite{kirillov2023segment} & \textcolor{gray}{ICCV'23} &
      62.54 & 76.86 & 79.11 & 32.00 & 45.66 & 52.60 \\
      SAM-Adapter~\cite{chen2023sam} & \textcolor{gray}{ICCVW'23} & 26.82 & 36.78 & 63.24 & \underline{44.40} & \underline{48.46} & 66.75 \\
      SAMCT~\cite{lin2024samct} & \textcolor{gray}{ICLR'24} & 34.50 & 43.80 & 49.19 & 43.55 & 46.62 & \textbf{90.50}  \\
      CAT-SAM~\cite{xiao2024cat} & \textcolor{gray}{ECCV'24} & 24.73 & 33.50 & 55.12 & 43.48 & 46.51 & 46.01 \\
      DenseSAM~\cite{zhou2025densesam} & \textcolor{gray}{IJCAI'25} & 29.18 & 40.78 & 74.18 & 37.59 & 48.36 & 49.94 \\
      MatSAM~\cite{li2025novel} & \textcolor{gray}{Acta Materialia'25} & \underline{62.73} & \underline{77.00} & \underline{79.36} & 33.04 & 47.15 & 54.34 \\
      $\mu$SAM~\cite{archit2025segment} & \textcolor{gray}{Nature Methods'25} &
      31.10 & 41.38 & 46.76 & 43.48 & 46.51 & 43.48 \\
      \rowcolor{blue!7}
      \textbf{SCoP-SAM} (Ours) & \textcolor{gray}{This Paper} &
      \textbf{67.26} & \textbf{80.42} & \textbf{80.43} & \textbf{57.25} & \textbf{67.48} & \underline{80.93} \\
      \bottomrule
    \end{tabular}
  }
  \caption{Comparison results against state-of-the-art methods on NBS and UHCS. The best and second-best results are in \textbf{bold} and \underline{underlined}.}
  \label{tab:compare-NBS-multi-datasets} 
\end{table*}

\subsection{Prior-aware Prompting Encoder (PPE)}

Upon obtaining $\mathbf{F}_{\text{SCP}}$, we embed it into the subsequent Prior-aware Prompting Encoder module to generate a robust prior prompt that guides downstream image encoding. Specifically, we introduce a learnable projection network $f_{\text{proj}}$ to project $\mathbf{F}_{\text{SCP}}$ onto common prompting tokens $\mathbf{F}_{\text{com}}$. Subsequently, we add $\mathbf{F}_{\text{com}}$ with $L_p$ learnable tokens $\mathbf{T}_{\text{lrn}}^i$ to generate prior-aware prompting tokens $\mathbf{T}_{\text{prio}}^i$ for feeding into the $i$-th transformer encoder $\mathcal{E}^{i}$. This process can be formulated as:
\begin{align}
\mathbf{F}_{\text{com}}&=f_{\text{proj}}(\mathbf{F}_{\text{SCP}}),\\
\mathbf{T}_{\text{prio}}^{i}&= \mathbf{F}_{\text{com}}\oplus\mathbf{T}_{\text{lrn}}^i,\ i=1,\cdots,N\!-\!1\\
\mathbf{T}_{\text{all}}^i&=\mathcal{E}^i(\text{Concat}(\mathbf{T}_{\text{im}}^i, \mathbf{T}_{\text{prio}}^{i})),\ i=1,\cdots,N\!-\!1,\\
\mathbf{T}_{\text{im}}^{N}&=\mathcal{E}^N(\mathbf{T}_{\text{all}}^{N-1}),
\end{align}
where $f_{\text{proj}}$ is a learnable network consisting of a convolutional layer, GELU, average pooling, and an MLP.
Notably, the PPE module incorporates $N$ transformer encoders $\{\mathcal{E}^{i}\}_{i=1}^N$, where $\mathcal{E}^{i+1}$ encodes the output of $\mathcal{E}^{i}$, updating the prior-aware prompting token input from $\mathbf{T}_{\text{prio}}^{i}$ to $\mathbf{T}_{\text{prio}}^{i+1}$ during encoding. For the final transformer encoder $\mathcal{E}^{N}$, we retain only $\mathbf{T}_{\text{im}}^{N}$ (i.e., the prior-enhanced image embedding tokens) and exclude $\mathbf{T}_{\text{prio}}^{N}$ from its output.

\begin{figure}[t]
\centering
\includegraphics[width=0.43\textwidth]{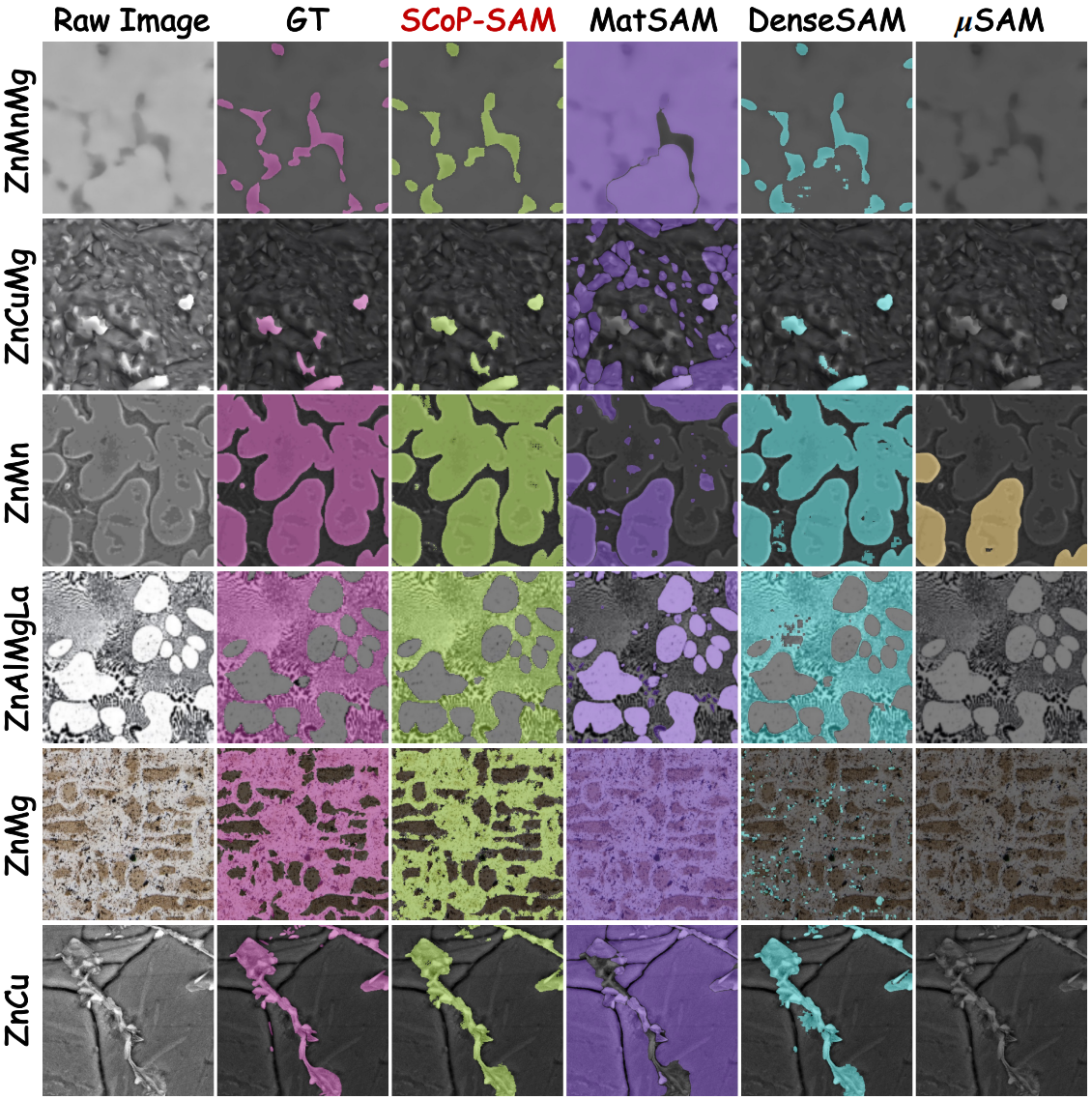}
\caption{Visualization of the proposed SCoP-SAM.
}
\label{visualization}
\end{figure}

\subsection{Prior-enhanced Mask Decoder (PMD)}

The PMD module integrates a spatial context prior and maps the prior-aware, token-guided image embedding to an output mask.
Specifically, after obtaining the image embedding $\mathbf{T}_{\text{im}}^{N}$, we add the spatial context prior $\mathbf{F}_{\text{SCP}}$ to it and feed their integration $\mathbf{F}_{\text{fuse}}$ into a two-block decoder $\mathcal{D}_1$. Each block of $\mathcal{D}_1$ contains a self-attention layer, a cross-attention (token-to-image) layer, an MLP, and a cross-attention (image-to-token) layer. A set of learnable output tokens $\mathbf{O}_{\text{self}}$ undergoes self-attention and serves as a query to attend to the prior-enhanced image embedding $\mathbf{F}_{\text{fuse}}$, enabling it to capture global image context. 
$\mathbf{F}_{\text{fuse}}$ acts as the key and value for the token-to-image cross-attention, and the refined output token $\mathbf{O}_{\text{self}}^{\text{re}}$ is updated via an MLP to integrate image contextual information. $\mathbf{F}_{\text{fuse}}$ serves as a query to attend to the updated output token via the image-to-token cross-attention, injecting global guidance into local spatial features and obtaining the prior-enhanced image embedding $\mathbf{F}_{\text{fuse}}^{\text{en}}$.
After a two-block decoder $\mathcal{D}_1$, $\mathbf{F}_{\text{fuse}}^{\text{en}}$ is upscaled via two transposed convolutions with GELU activations and layer normalization, yielding a downscaled feature map $\mathbf{F}_{\text{down}}$ relative to the input image.
The refined output token $\mathbf{O}_{\text{self}}^{\text{re}}$ is fed into a token-to-image cross-attention layer, followed by an MLP that produces an upscaled image embedding $\mathbf{F}_{\text{fuse}}^{\text{up}}$.
The point-wise sum of $\mathbf{F}_{\text{fuse}}^{\text{up}}$ and $\mathbf{F}_{\text{com}}$ (i.e., $f_{\text{com}}(\mathbf{F}_{\text{SCP}})$) followed by the point-wise multiplication of $\mathbf{F}_{\text{down}}$ and $\mathbf{F}_{\text{com}}$ yields the foreground probability for each pixel, producing the final segmentation mask.

\subsection{Training}
The overall training objective of SCoP-SAM is formulated as a linearly weighted multi-task loss function that integrates three functionally complementary core loss components: a region-overlap optimization component (Dice loss), a pixel-wise binary classification component (binary cross-entropy loss), and an auxiliary intersection-over-union (IoU) regression component for segmentation consistency.

\section{Experiments}

\subsection{Experimental Setup}
\noindent\textbf{Datasets.}
To evaluate the proposed SCoP-SAM, we conduct experiments on various datasets, including IPSM-Bench, NBS~\cite{stuckner2022microstructure}, and UHCS~\cite{na2023unified}. 
All methods are trained on the IPSM-Bench training set and evaluated on the test sets of IPSM-Bench, NBS, and UHCS.
This design is due to the training sets for NBS and UHCS being too small to support model training.
Specifically, for IPSM-Bench, the training set comprises 800 samples: 246 from the OM modality and 554 from the SEM modality, while the test set comprises 254 samples: 132 from the OM modality and 122 from the SEM modality. More details are provided in Figure~\ref{statistics} and subsection~\ref{subsection:Statistics}.
UHCS has 5 labeled SEM images of cementite particles for testing.
NBS has 9 labeled SEM images of nisuperalloy for testing.

\noindent\textbf{Metrics.}
Following prior efforts~\cite{zhou2025densesam,li2025novel,chen2023sam}, we adopt three pixel-level metrics: Intersection over Union (IoU), F1 score, and Precision. Specifically, we compute the performance for the foreground and background classes separately for each metric and report the mean across two classes, denoted mIoU, mF1, and mPre.

\noindent\textbf{Implementation Details.}
To ensure the fairness of comparative experiments, all experiments adhere to consistent configurations: 1) all methods use SAM ViT-H~\cite{kirillov2023segment} as the backbone; 2) all methods use their official open-source implementations and settings, trained from scratch on IPSM-Bench's training set. We train SCoP-SAM on four NVIDIA A800 GPUs with a batch size of 4 for 50 epochs using the AdamW optimizer and a learning rate of 1e-4.

\subsection{Results and Analysis}

\noindent\textbf{Performance Comparisons on IPSM-Bench.}
We compare our SCoP-SAM with SOTA methods~\cite{zhou2025densesam,xiao2024cat,chen2023sam,lin2024samct,archit2025segment,kirillov2023segment,li2025novel} on IPSM-Bench.
In Table~\ref{tab:compare-IPSM-Bench-multi-datasets}, SCoP-SAM consistently outperforms all methods. For instance, SCoP-SAM improves mIoU by 11.70\% over DenseSAM, 9.51\% over SAMCT, and 28.47\% over CAT-SAM. We attribute these improvements primarily to incorporating a spatial context prior and prior-aware prompts into the learnable modules of both the encoder and decoder, enabling robust performance under the challenges of intermediate phase segmentation, including low inter-phase contrast, blurred or incomplete boundaries, a high proportion of small-scale intermediate phases, heterogeneous morphologies, and interference from noise.

\noindent\textbf{Performance Comparisons on NBS and UHCS.}
We further investigate the generalization of SCoP-SAM in few-shot settings with pronounced distribution shifts in materials data, such as large variations in precipitate size and more drastic morphological changes. To this end, we evaluate our SCoP-SAM on the more challenging NBS dataset and report the results in Table \ref{tab:compare-NBS-multi-datasets}. Notably, the original SAM already achieves 62.54\% on NBS, whereas several SAM-based variants (i.e., SAM-Adapter, SAMCT, CAT-SAM, DenseSAM, and $\mu$SAM) exhibit performance degradation, suggesting that their generalization under severe domain shifts remains limited. In contrast, SCoP-SAM outperforms MatSAM by 4.53\%, further demonstrating the robustness of our method beyond the training data distribution. We attribute this advantage primarily to SCoP-SAM's use of multi-source spatial context priors to better exploit structural cues, enabling more stable object localization and boundary delineation under large-scale variations and morphological heterogeneity, thereby mitigating performance drops caused by few-shot learning and cross-domain distribution discrepancies.
For the UHCS dataset, SCoP-SAM outperforms DenseSAM, SAM-Adapter, and SAMCT by 19.66\%, 12.85\%, and 13.70\% in mIoU, respectively, and demonstrates strong generalization, indicating that our IPSM-Bench also covers diverse intermediate phase morphologies and imaging conditions with high representativeness.
Notably, SAMCT achieves a higher mPre (90.50\%) than SCoP-SAM (80.93\%) on UHCS; yet it fails to achieve a higher mIoU. This is because SAMCT has a higher miss rate (i.e., false negatives), predicting only a small set of high-confidence regions, which inflates precision, but insufficient mask coverage substantially degrades IoU.

\begin{table}[t]
  \centering
  \adjustbox{width=\linewidth}{
    \setlength{\tabcolsep}{4pt}
    \begin{tabular}{lccc}
      \toprule
      Module & mIoU (\%) & mF1 (\%) & mPre (\%) \\
      \midrule      Baseline (ViT-H)       & 29.05 & 44.83 & 60.98 \\
      Baseline+SCP        & 48.18 & 61.72 & 69.98 \\
      Baseline+SCP+PPE      & \underline{59.49} & \underline{72.98} & \underline{81.38} \\
      \rowcolor{blue!7}\textbf{Baseline+SCP+PPE+PMD}   & \textbf{71.52} & \textbf{82.82} & \textbf{87.55} \\
      \bottomrule
    \end{tabular}
  }
  \caption{Ablation study results on IPSM-Bench. Our strategy is highlighted in \textbf{bold}.}
  \label{tab:ablation-modules}
\end{table}

\subsection{Ablation Study}
We evaluate the effects of three core modules (i.e., SCP, PPE, and PMD). In Table~\ref{tab:ablation-modules}, SCP alone achieves 48.18\% mIoU; adding PPE improves performance to 59.49\%; integrating PMD into the complete model yields optimal performance of 71.52\%. These results demonstrate that each module positively contributes to segmentation performance and that the modules work synergistically to enhance overall performance, supporting SCoP-SAM's performance advantage.

\subsection{Visualization Analysis}
We provide a visual analysis of SEM and OM images of six typical zinc alloys, as shown in Figure~\ref{visualization}. The SEM and OM images present distinct challenges: SEM images suffer from low contrast and background noise, while OM images exhibit complex topological structures of microstructural phases.
Compared with the problems of over-segmentation, under-segmentation, or even complete failure in MatSAM, DenseSAM, and $\mu$SAM, our SCoP-SAM can stably and accurately restore irregular morphology and detailed boundaries, achieving high consistency with ground-truth (GT) annotations across all samples and demonstrating excellent cross-material and cross-modality generalization and robustness.

\section{Conclusion}
In this paper, we present IPSM-Bench, a new, large-scale, diverse, and high-quality benchmark for intermediate phase segmentation in zinc alloy microstructures, providing a reliable foundation for training and testing segmentation models. We further propose SCoP-SAM, which incorporates spatial context priors into image encoding and mask decoding generation, enabling more reliable use of the gradient structure and grayscale properties while improving both segmentation performance and boundary quality. Extensive experiments demonstrate that SCoP-SAM achieves state-of-the-art results on IPSM-Bench and delivers clear gains on two other public alloy benchmarks, indicating strong segmentation ability across various alloy scenarios.

\bibliographystyle{named}
\bibliography{ijcai26}

@inproceedings{wugausam,
  title={GauSAM: Contour-Guided 2D Gaussian Fields for Multi-Scale Medical Image Segmentation with Segment Anything},
  author={Wu, Jinxuan and Wang, Jiange and Zhang, Dongdong},
  booktitle={The Thirty-ninth Annual Conference on Neural Information Processing Systems},
  year={2025}
}

@article{marks2025cellsam,
  title={CellSAM: a foundation model for cell segmentation},
  author={Marks, Markus and Israel, Uriah and Dilip, Rohit and Li, Qilin and Yu, Changhua and Laubscher, Emily and Iqbal, Ahamed and Pradhan, Elora and Ates, Ada and Abt, Martin and others},
  journal={Nature Methods},
  pages={1--9},
  year={2025}
}

@article{ma2024segment,
  title={Segment anything in medical images},
  author={Ma, Jun and He, Yuting and Li, Feifei and Han, Lin and You, Chenyu and Wang, Bo},
  journal={Nature Communications},
  volume={15},
  number={1},
  pages={654},
  year={2024}
}

@article{archit2025segment,
  title={Segment anything for microscopy},
  author={Archit, Anwai and Freckmann, Luca and Nair, Sushmita and Khalid, Nabeel and Hilt, Paul and Rajashekar, Vikas and Freitag, Marei and Teuber, Carolin and Spitzner, Melanie and Tapia Contreras, Constanza and others},
  journal={Nature Methods},
  volume={22},
  number={3},
  pages={579--591},
  year={2025}
}

@inproceedings{chen2024robustsam,
  title={Robustsam: Segment anything robustly on degraded images},
  author={Chen, Wei-Ting and Vong, Yu-Jiet and Kuo, Sy-Yen and Ma, Sizhou and Wang, Jian},
  booktitle={Proceedings of the IEEE/CVF Conference on Computer Vision and Pattern Recognition},
  pages={4081--4091},
  year={2024}
}

@inproceedings{chen2023sam,
  title={Sam-adapter: Adapting segment anything in underperformed scenes},
  author={Chen, Tianrun and Zhu, Lanyun and Deng, Chaotao and Cao, Runlong and Wang, Yan and Zhang, Shangzhan and Li, Zejian and Sun, Lingyun and Zang, Ying and Mao, Papa},
  booktitle={Proceedings of the IEEE/CVF International Conference on Computer Vision},
  pages={3367--3375},
  year={2023}
}

@inproceedings{zhou2025densesam,
  title={DenseSAM: Semantic Enhance SAM For Efficient Dense Object Segmentation},
  author={Zhou, Linyun and Hu, Jiacong and Zhang, Shengxuming and Du, Xiangtong and Song, Mingli and Zhang, Xiuming and Feng, Zunlei},
  booktitle={Proceedings of the Thirty-Fourth International Joint Conference on Artificial Intelligence},
  pages={7994--8002},
  year={2025}
}

@inproceedings{xiao2024cat,
  title={Cat-sam: Conditional tuning for few-shot adaptation of segment anything model},
  author={Xiao, Aoran and Xuan, Weihao and Qi, Heli and Xing, Yun and Ren, Ruijie and Zhang, Xiaoqin and Shao, Ling and Lu, Shijian},
  booktitle={European Conference on Computer Vision},
  pages={189--206},
  year={2024}
}

@inproceedings{kirillov2023segment,
  title={Segment anything},
  author={Kirillov, Alexander and Mintun, Eric and Ravi, Nikhila and Mao, Hanzi and Rolland, Chloe and Gustafson, Laura and Xiao, Tete and Whitehead, Spencer and Berg, Alexander C and Lo, Wan-Yen and others},
  booktitle={Proceedings of the IEEE/CVF International Conference on Computer Vision},
  pages={4015--4026},
  year={2023}
}

@article{ma2025alloy,
  title={Alloy microstructure segmentation through SAM and domain knowledge without extra training},
  author={Ma, Xudong and Zhang, Yuqi and Wang, Chenchong and Xu, Wei},
  journal={Scripta Materialia},
  volume={260},
  pages={116581},
  year={2025}
}

@article{abebe2025sam,
  title={SAM-I-Am: Semantic boosting for zero-shot atomic-scale electron micrograph segmentation},
  author={Abebe, Waqwoya and Strube, Jan and Guo, Luanzheng and Tallent, Nathan R and Bel, Oceane and Spurgeon, Steven and Doty, Christina and Jannesari, Ali},
  journal={Computational Materials Science},
  volume={246},
  pages={113400},
  year={2025}
}

@article{li2025novel,
  title={A novel training-free approach to efficiently extracting material microstructures via visual large model},
  author={Li, Changtai and Han, Xu and Yao, Chao and Guo, Yu and Li, Zixin and Jiang, Lei and Liu, Wei and Huang, Haiyou and Fu, Huadong and Ban, Xiaojuan},
  journal={Acta Materialia},
  volume={290},
  pages={120962},
  year={2025}
}

@article{guru2025machine,
  title={Machine learning pipeline for Structure-Property modeling in Mg-alloys using microstructure and texture descriptors},
  author={Guru, Mahish K and Bohlen, Jan and Aydin, Roland C and Khalifa, Noomane Ben},
  journal={Acta Materialia},
  pages={121132},
  year={2025}
}

@article{huang2025additively,
  title={Additively manufactured biodegradable Zn-Mn-based implants with an unprecedented balance of strength and ductility},
  author={Huang, Chengcong and Wang, Yizhu and Yang, Fan and Shi, Yixuan and Zhao, Shangyan and Li, Xuan and Lu, Yuchen and Wu, Yuzhi and Zhou, Jie and Zadpoor, Amir A and others},
  journal={Acta Biomaterialia},
  volume={196},
  pages={506--522},
  year={2025}
}

@article{yang2023lithium,
  title={Lithium-induced optimization mechanism for an ultrathin-strut biodegradable Zn-based vascular scaffold},
  author={Yang, Hongtao and Jin, Dawei and Rao, Jiancun and Shi, Jiahui and Li, Guannan and Wang, Cheng and Yan, Kai and Bai, Jing and Bao, Guo and Yin, Meng and others},
  journal={Advanced Materials},
  volume={35},
  number={19},
  pages={2301074},
  year={2023}
}

@article{tong2023znp,
  title={ZnP-Coated Zn-1Cu-0.1Ti Membrane with High Strength-Ductility, Antibacterial Ability, Cytocompatibility, and Osteogenesis for Biodegradable Guided Bone Regeneration Applications},
  author={Tong, Xian and Han, Yue and Zhu, Li and Zhou, Runqi and Lin, Zhiqiang and Wang, Hongning and Huang, Shengbin and Li, Yuncang and Ma, Jianfeng and Wen, Cuie and others},
  journal={Advanced Functional Materials},
  volume={33},
  number={31},
  pages={2214657},
  year={2023}
}

@article{stuckner2022microstructure,
  title={Microstructure segmentation with deep learning encoders pre-trained on a large microscopy dataset},
  author={Stuckner, Joshua and Harder, Bryan and Smith, Timothy M},
  journal={npj Computational Materials},
  volume={8},
  number={1},
  pages={200},
  year={2022}
}

@article{decost2017uhcsdb,
  title={UHCSDB: ultrahigh carbon steel micrograph database: tools for exploring large heterogeneous microstructure datasets},
  author={DeCost, Brian L and Hecht, Matthew D and Francis, Toby and Webler, Bryan A and Picard, Yoosuf N and Holm, Elizabeth A},
  journal={Integrating Materials and Manufacturing Innovation},
  volume={6},
  number={2},
  pages={197--205},
  year={2017}
}

@article{li2020additively,
  title={Additively manufactured biodegradable porous metals},
  author={Li, Yageng and Jahr, Holger and Zhou, Jie and Zadpoor, Amir Abbas},
  journal={Acta Biomaterialia},
  volume={115},
  pages={29--50},
  year={2020}
}

@article{luengo2022tutorial,
  title={A tutorial on the segmentation of metallographic images: Taxonomy, new MetalDAM dataset, deep learning-based ensemble model, experimental analysis and challenges},
  author={Luengo, Julian and Moreno, Raul and Sevillano, Ivan and Charte, David and Pelaez-Vegas, Adrian and Fernandez-Moreno, Marta and Mesejo, Pablo and Herrera, Francisco},
  journal={Information Fusion},
  volume={78},
  pages={232--253},
  year={2022}
}

@article{decost2019high,
  title={High throughput quantitative metallography for complex microstructures using deep learning: A case study in ultrahigh carbon steel},
  author={DeCost, Brian L and Lei, Bo and Francis, Toby and Holm, Elizabeth A},
  journal={Microscopy and Microanalysis},
  volume={25},
  number={1},
  pages={21--29},
  year={2019}
}

@article{na2023unified,
  title={A unified microstructure segmentation approach via human-in-the-loop machine learning},
  author={Na, Juwon and Kim, Se-Jong and Kim, Heekyu and Kang, Seong-Hoon and Lee, Seungchul},
  journal={Acta Materialia},
  volume={255},
  pages={119086},
  year={2023}
}

@article{ma2023training,
  title={Training tricks for steel microstructure segmentation with deep learning},
  author={Ma, Xudong and Yu, Yunhe},
  journal={Processes},
  volume={11},
  number={12},
  pages={3298},
  year={2023}
}

@ARTICLE{lin2024samct,
  author={Lin, Xian and Xiang, Yangyang and Wang, Zhehao and Cheng, Kwang-Ting and Yan, Zengqiang and Yu, Li},
  journal={IEEE Transactions on Medical Imaging}, 
  title={SAMCT: Segment Any CT Allowing Labor-Free Task-Indicator Prompts}, 
  year={2025},
  volume={44},
  number={3},
  pages={1386--1399}
}

@article{wu2026hierarchical,
  title={Hierarchical Open-vocabulary Part-object Segmentation with Knowledge-guided SAM},
  author={Wu, Xin-Jian and Liu, Cheng-Lin},
  journal={Machine Intelligence Research},
  volume={23},
  number={1},
  pages={214--226},
  year={2026}
}

@article {ji2024segment,
author = {Ji, Wei and Li, Jingjing and Bi, Qi and Liu, Tingwei and Li, Wenbo and Cheng, Li},
journal = {Machine Intelligence Research},
title = {Segment Anything Is Not Always Perfect: An Investigation of SAM on Different Real-world Applications},
year = {2024},
volume = {21},
number = {4},
pages = {617-630}
}

\end{document}